\documentclass[a4paper,fleqn]{cas-dc}

\usepackage[authoryear]{natbib}

\usepackage{todonotes}
\usepackage{longtable}
\usepackage{algorithm,algorithmic}

\newcommand{\cs}[1]{\texttt{\symbol{`\\}#1}}
\usepackage{subcaption}
\usepackage[normalem]{ulem}
\usepackage{float} 
\begin{document}
\let\WriteBookmarks\relax
\def\floatpagepagefraction{1}
\def\textpagefraction{.001}
\shorttitle{Early Diagnosis AF Recurrence}
\shortauthors{A.G. Domingo-Aldama et~al.}

\title[mode = title]{Early Diagnosis of Atrial Fibrillation Recurrence: A Large Tabular Model Approach with Structured and Unstructured clinical data}

\author[1]{Ane G. {Domingo-Aldama}}[orcid=0009-0000-8202-1099]
\ead{ane.garciad@ehu.eus}

\author[1]{Marcos {Merino Prado}}[orcid=0009-0004-9610-322X]
\ead{mmerino08p@gmail.com}

\author[2]{Alain {García Olea}}[orcid=0000-0002-8742-0600]
\ead{alain.garciaolea@osakidetza.eus}

\author[1]{Koldo {Gojenola Galletebeitia}}[orcid=0000-0002-2116-6611]
\ead{koldo.gojenola@ehu.eus}

\author[1]{Josu {Goikoetxea Salutregi}}[orcid=0000-0001-5568-4014]
\ead{josu.goikoetxea@ehu.eus}

\author[1]{Aitziber {Atutxa Salazar}}[orcid=0000-0003-4512-8633]
\ead{aitziber.atucha@ehu.eus}

\affiliation[1]{organization={University of the Basque Country}, 
                city={Bilbao},
                state={Biscay},
                country={Spain}}

\affiliation[2]{organization={Basurto University Hospital}, 
                city={Bilbao},
                state={Biscay},
                country={Spain}}

\begin{abstract} % not more than 250 words
BACKGROUND: Atrial fibrillation (AF), the most common arrhythmia, is linked to high morbidity and mortality. In a fast-evolving AF rhythm control treatment era, predicting AF recurrence after its onset may be crucial to achieve the optimal therapeutic approach, yet traditional scores like CHADS2-VASc, HATCH, and APPLE show limited predictive accuracy. Moreover, early diagnosis studies often rely on codified electronic health record (EHR) data, which may contain errors and missing information. \\

OBJECTIVE: This study aims to predict AF recurrence between one month and two years after onset by evaluating traditional clinical scores, ML models, and our LTM approach. Moreover, another objective is to develop a methodology for integrating structured and unstructured data to enhance tabular dataset quality. \\

METHODS: A tabular dataset was generated by combining structured clinical data with free-text discharge reports processed through natural language processing techniques, reducing errors and annotation effort. A total of 1,508 patients with documented AF onset were identified, and models were evaluated on a manually annotated test set. The proposed approach includes a LTM compared against traditional clinical scores and ML models.\\

RESULTS: The proposed LTM approach achieved the highest predictive performance, surpassing both traditional clinical scores and ML models. Additionally, the gender and age bias analyses revealed demographic disparities. \\

CONCLUSION: The integration of structured data and free-text sources resulted in a high-quality dataset. The findings emphasize the limitations of traditional clinical scores in predicting AF recurrence and highlight the potential of ML-based approaches, particularly our LTM model.
\end{abstract}

\begin{graphicalabstract}
\includegraphics{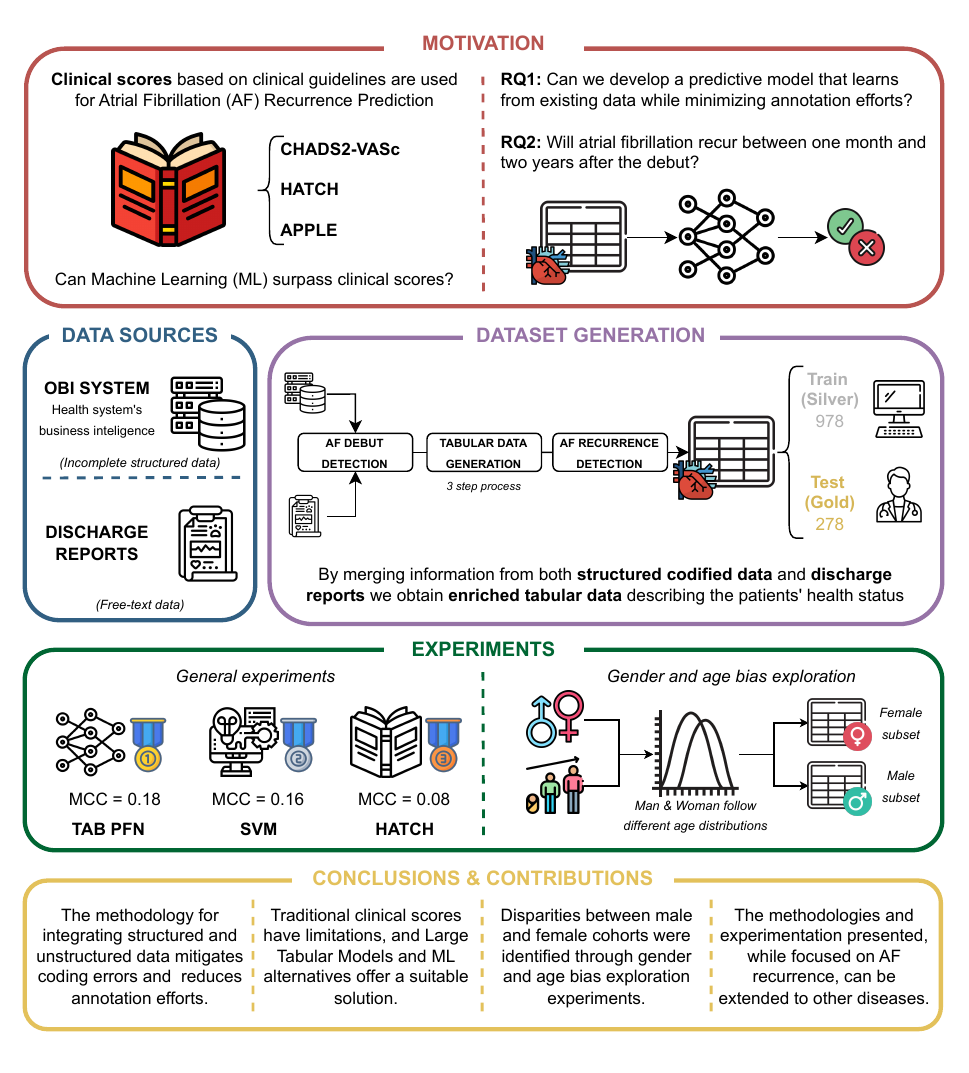}
\end{graphicalabstract}

\begin{highlights} % not more than 6 85 characters
\item Dataset combines structured EHR data with unstructured reports to improve quality.

\item A three-step NLP pipeline extracts tabular data from discharge reports accurately.

\item The study predicts AF recurrence within 1 month to 2 years after AF onset.

\item Traditional clinical scores are compared to SVM and LTM for predictive performance.

\item LTM with minimal data preprocessing, achieves the best results consistently.

\item A gender and age bias analysis evaluated model performance across populations.
\end{highlights}

\begin{keywords} % not more than 6
Atrial Fibrillation Recurrence; Electronic Health Records; Early Diagnosis; Clinical Scores; Large Tabular Models; Natural Language Processing
\end{keywords}

\maketitle

\section{Introduction}\label{sec:intro}   

Early detection (ED) of health risks has become a critical focus in modern medical research. The ability to recognize potential health threats at their earliest stages allows for timely intervention, significantly improving patient outcomes and reducing the burden of late-stage disease management.

The integration of Artificial Intelligence (AI) in biomedicine has accelerated ED \citep{Kumar2023}, improved diagnostic accuracy \citep{kanjee2023Jama} and advanced predictive analytics and personalized medicine. The power of AI-driven algorithms lies on uncovering subtle patterns in medical data, enhancing proactive healthcare and shifting the focus from reactive treatment to prevention while optimizing healthcare resources \citep{yelne2023harnessing, goh2021artificial}.

Given the anticipated rise in cardiovascular diseases (CVDs) as a leading cause of morbidity and mortality worldwide, particularly in developing nations \citep{celermajer2012cardiovascular, lippi2021global}, research into the ED of CVDs has become increasingly important. Within this context, the early identification of atrial fibrillation (AF)—the most common type of arrhythmia— and its recurrence has garnered significant attention due to its potential to enable timely therapeutic interventions and mitigate the risk of severe complications. This underscores the critical role of AF detection in cardiology, particularly in reducing the overall burden of CVDs.

Currently, existing clinical scoring systems, such as CHADS2-VASc, HATCH, and APPLE, are widely employed to predict new-onset AF and postoperative AF recurrence \citep{ramos2023use, vitali2019cha2ds2, huang2024better}. 

Our main goal is to improve and go beyond the results provided by the mentioned scores with a minimal annotation effort overcoming the challenges faced when using real medical data (i.e missing values, lack of standardization, errors). To achieve this, we propose a methodology that, while applied to the ED of AF recurrence, is generalizable to any other pathology. Our approach leverages Large Tabular Models (LTMs), utilizing tabular data from structured electronic health records (EHRs) and enriching it with insights extracted from free-text clinical notes, thereby enhancing the predictive performance of these models. 

Large Tabular Models (LTMs) are a specialized class of AI models designed to process and generate insights from structured tabular data.

By combining both structured and unstructured data sources, the proposed approach seeks to overcome the shortcomings of systems that rely exclusively on structured EHR data—data that is typically housed within a healthcare institution’s Business Intelligence system or database. First, structured data is produced through a coding process in which documentation services convert information from clinical notes into coded entries. This process is susceptible to human error, which can significantly impact its accuracy and reliability. As highlighted by \cite{garcia2021role}, error rates in identifying atrial fibrillation (AF) onset can reach as high as 26\%. Similarly, \cite{botsis2010secondary} report that missing information in datasets related to endocrine pancreatic tumors varies widely, with rates ranging from 6\% to 46\% of variables. Second, despite extensive efforts to establish coding standards such as ICD10, OPCS, and SNOMED, no universal guidelines exist regarding the extent of coding in clinical records. Practices vary widely—some countries or institutions document only the primary diagnosis and treatment, while others record all final diagnoses. In some cases, diagnoses that are not directly observed by the clinician, but emerge from exploratory tests, are also included depending on the institution. In some other cases, when there is no clear diagnosis, some health systems code relevant symptoms (NHSGuideLine, MedicareGuideLine). 

While prior studies have primarily focused on predicting the first occurrence of AF \citep{tseng2021prediction} or predicting AF recurrence 1 year after catheter ablation, this research introduces a new perspective by focusing on AF recurrence and within a previously unexplored time frame—from one month to two years after its initial onset. The extension of the time-frame from one to two years aligns seamlessly with the ED philosophy, emphasizing the value of proactive early intervention by enabling forecasts two years ahead unlike other previous studies limited to one year ahead.

Building on these insights, the core research questions explored in this study are two:

\begin{itemize}
    \item \textbf{RQ1: Can we develop a predictive model that learns from existing data while minimizing annotation efforts?} Given that clinicians' time is highly valuable and data annotation is a labor-intensive process, it is crucial to establish a methodology that effectively reduces this burden.
    \item \textbf{RQ2: Will atrial fibrillation recur within the next two years?}
    From a clinical standpoint, this question is essential for guiding the decision-making process of electrophysiologists and arrhythmologists, helping them tailor treatment strategies and optimize patient care. Although symptomatic young patients are normally directed to rhythm control strategies, in many middle-aged and older patients both rhythm control and rate control could be indicated, so interventions aimed to avoid the AF recurrence are usually delayed or avoided. The AF recurrence prediction could lead to a more personalized clinician-based decision 
    
    \end{itemize}

\begin{figure*}[!htb]
    \centering
    % First subfigure
    \begin{subfigure}{\textwidth}
        \centering
        \includegraphics[width=\textwidth]{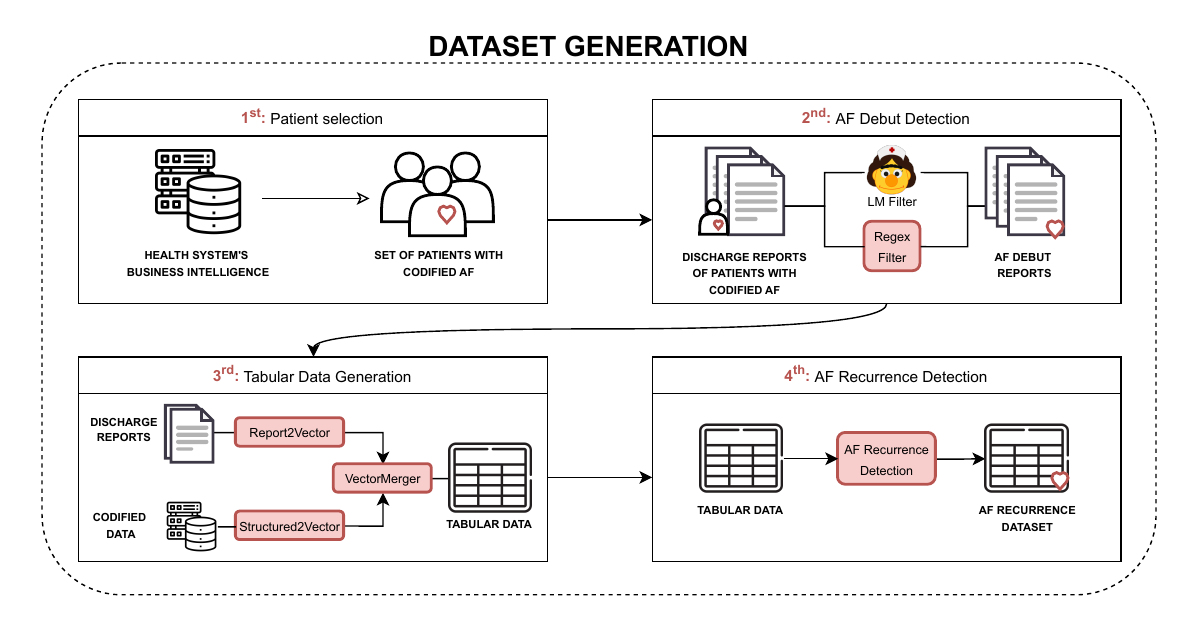}
        \caption{Summary of the dataset generation process.}
        \label{fig:architecture_subfigure1}
    \end{subfigure}
    
    \vspace{0.5cm} % Space between subfigures

    % Third subfigure
    \begin{subfigure}{\textwidth}
        \centering
        \includegraphics[width=\textwidth]{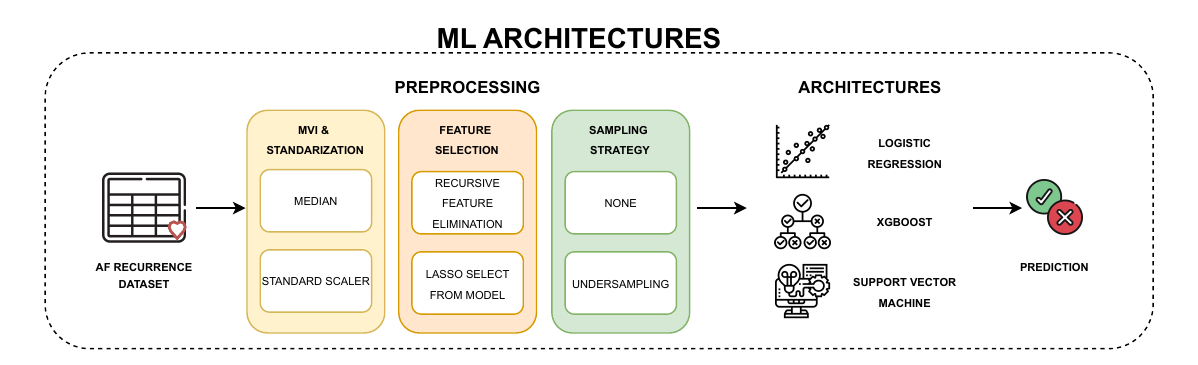}
        \caption{Summary of the ML models and preprocessing techniques used for the experimentation.}
        \label{fig:architecture_subfigure2}
    \end{subfigure}

    \vspace{0.5cm} % Space between subfigures

    % Second subfigure
    \begin{subfigure}{\textwidth}
        \centering
        \includegraphics[width=\textwidth]{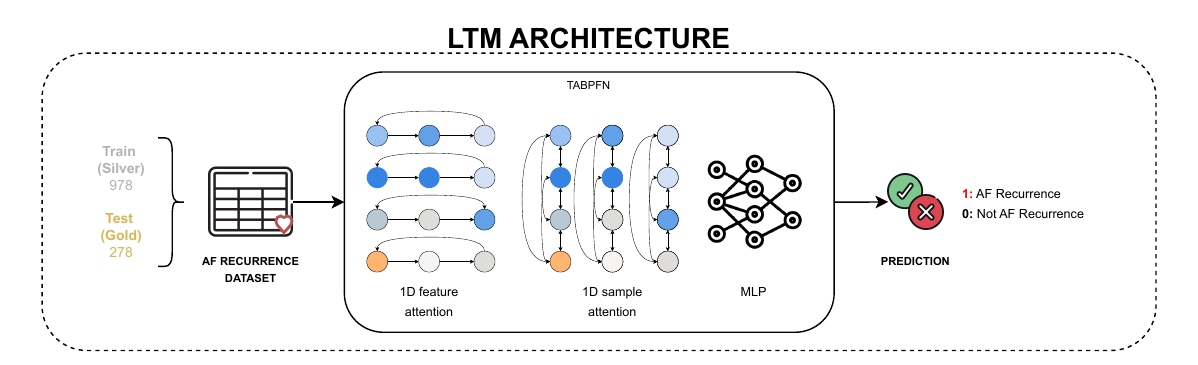}
        \caption{Overview of the Large Tabular Model used for the experimentation.}
        \label{fig:architecture_subfigure3}
    \end{subfigure}
            
    \caption{Overview of the generated dataset and the experimental setup for AF recurrence prediction.}
    \label{fig:architecture}
\end{figure*}

After addressing these research questions, we outline the key contributions of our study:
\begin{itemize}
\item A methodology for integrating structured and unstructured data that minimizes the impact of coding errors, ensures more comprehensive information, and reduces annotation effort (see \autoref{fig:architecture_subfigure1}). This integration is crucial in the two steps of the dataset generation.
 
\item Exploration of traditional ML approaches that require data pre-processing and feature engineering (see \autoref{fig:architecture_subfigure2}) in comparison to LTM approaches with no such requirements (see \autoref{fig:architecture_subfigure3}), simplifying the learning process.

\item   Comparison of Clinical Scoring Systems with ML techniques and LTMs: to the best of our knowledge, this is the first study to compare the mentioned clinical scoring systems with traditional ML methods and LTMs.

\item Expansion of the ED's temporal coverage: This work aims to predict AF recurrence within a two years time-span.

\item Gender and age bias exploration: This study analyzes the impact of gender and age on AF recurrence prediction by evaluating model performance across separate male and female cohorts and conducting experiments on younger patients. The results highlight significant demographic disparities across all techniques.

\end{itemize}

\section{Related Work} \label{sec:rel_work}

Cardiological disease prediction \citep{naser2024review, bhatt2023effective, soni2011predictive} has emerged as a critical area of research in recent years, owing to its potential to enhance ED and management. AF, the most prevalent arrhythmia globally, holds particular importance in cardiology, especially concerning its recurrence. ED of AF recurrence enables timely interventions, thereby reducing the risk of severe complications such as strokes, heart failure, and dementia. Consequently, numerous AI-based methods for AF detection and prediction have been proposed in recent literature.

Research in this domain typically focuses on two primary scenarios: the prediction of new-onset AF and the recurrence of AF following therapeutic interventions. AI-driven models have demonstrated remarkable success in predicting incident AF, often outperforming conventional methods \citep{siontis2020will}. These studies leverage diverse datasets, including clinical data, cardiac imaging data, and electrophysiological data \citep{tseng2021prediction}. For instance, several studies have utilized EHRs to extract risk factors from large patient cohorts to predict the first episode of AF \citep{nadarajah2021predicting, hulme2019development, tiwari2020assessment}. Notably, \cite{sung2022automated} developed a ML model to predict the risk of newly detected atrial fibrillation post-stroke, incorporating both structured variables and unstructured clinical text processed through NLP. Their best-performing model surpassed traditional risk scores such as AS5F and CHASE-LESS. While this study shares similarities with our work, it relies on manually annotated corpora for training. In contrast, our research generates and employs a silver training dataset reducing the need for manual annotations. Furthermore, their approach is confined to a one-year prediction period, whereas our method extends the prediction horizon to two years.

A substantial portion of the existing literature is dedicated to predicting AF recurrence after catheter ablation, a widely used therapeutic approach to restore sinus rhythm \citep{packer2019effect}. As highlighted in the review by \cite{fan2023predictive}, ML-based methods demonstrate high performance in predicting AF recurrence after ablation, positioning them as a competitive and cost-effective alternative for post-ablation screening. The present study distinguishes itself from the aforementioned works by extending beyond the prediction of AF recurrence following ablation.

Many investigations incorporate clinical variables along with left atrial information obtained from echocardiograms \citep{knecht2024machine, zhou2022deep}. Others integrate clinical data with electrocardiogram findings \citep{qiu2024deep} or computed tomography images \citep{liu2024use, brahier2023using}, demonstrating improvements over traditional risk scores.
As mentioned in the introduction, our system is based exclusively on structured data combined with unstructured information derived from clinical notes. Availability of other sources of information is highly dependent on the specific health system. As noted in \cite{sung2022automated} and corroborated by our experience, these exploratory tests (e.g. ECGs, echocardiograms) are rarely available for all patients, as they are sometimes performed in private institutions outside the hospital setting or just for storage reasons. However, clinicians document the results of these tests in clinical notes using natural language, sometimes gathered through patient anamnesis. Unfortunately, often the outcome of the exploratory procedures themselves are not disclosed. Consequently, we sought to maintain experimental homogeneity and simplicity by exclusively employing the unstructured data contained in clinical notes—which are universally available for all patients—to evaluate whether this approach can surpass the performance of existing clinical scores.

The quality of EHR data is a critical factor in ensuring the reliability and accuracy of predictive models, particularly in the context of medical research. Numerous studies have addressed the common challenges associated with EHR data quality, as well as methods for assessing and improving it \citep{cruz2009data, lewis2023electronic, feder2018data, terry2019basic}. For example, \cite{jetley2019electronic} suggest using information extraction techniques or manual review of clinical notes to address gaps in structured data. This approach aligns with the hypothesis of the present project, which aims to utilize similar techniques for extracting relevant information from free-text discharge reports to complement, improve, and double check the quality of structured EHR data. 

The emergence of Large Language Models (LLMs) has significantly advanced ED, particularly in leveraging free-text clinical reports for predictive modeling \citep{zhou2024large}. While LLMs have demonstrated remarkable capabilities in processing unstructured medical text, a substantial portion of clinical information remains stored as structured tabular data, necessitating specialized approaches for effective utilization \citep{van2024tabular}. In this context, LTMs, such as TabPFN \citep{hollmann2025accurate}, offer a promising alternative. TabPFN has been shown to perform well in low-data settings and effectively handle high-dimensional data with missing values, challenges commonly encountered in medical datasets.

Despite all the advances, a significant gap remains in the prediction of AF recurrence following an initial event. This study aims to address this gap by proposing a methodology that integrates structured codified EHR data with information extracted from free text clinical reports to create high-quality patient cohorts and a silver training dataset. By doing so, we seek to enhance the accuracy and applicability of AF recurrence prediction, ultimately contributing to improved patient care and outcomes.

\section{Dataset} \label{sec:resources}

The present research has been possible thanks to the following three resources kindly provided by the Basque Public Healthcare System (Osakidetza). This dataset exclusively comprises clinical information from patients treated at Basurto Hospital in the Basque Country:

\begin{itemize}
    \item \textit{Discharge reports in Spanish}: A pool of $1.2\times10^6$ discharge reports dating from 2015 to 2020.
    \item \textit{Codified structured data from the Osakidetza Business Intelligence (OBI) system}: Clinical data for each patient, encoded by healthcare professionals using standardized coding systems and stored in a Business Intelligence platform.

    \item \textit{A minimal manual annotation of AF recurrence}: Annotations of the recurrence status of patients annotated in the cardiology department of BioBizkaia, used to label the test dataset.
\end{itemize}

\begin{figure*}[ht]
    \centering
    \includegraphics[width=17cm]{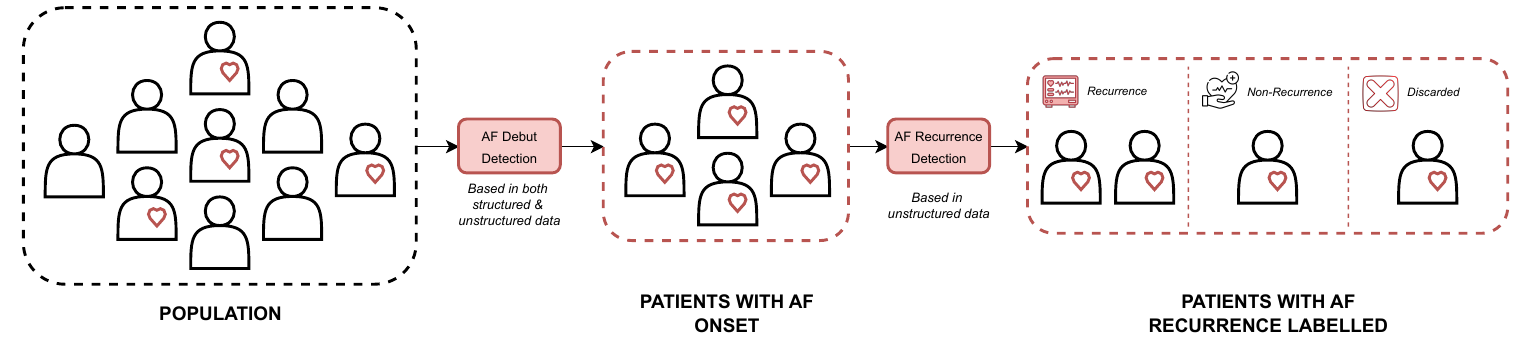}
    \caption{Summary of the AF recurrence cohort generation process. First, patients with an AF onset are identified, followed by the assignment of the AF recurrence label.}
    \label{fig:cohort}
\end{figure*}

The following subsections outline the three-step methodology proposed  for the the \textit{AF recurrence} cohort generation based on the data already mentioned (Discharge reports, Codified data, Manually annotated data) (see \autoref{fig:cohort}). First, we introduce details on the detection of AF onset episodes, including the patient inclusion criteria (\autoref{sec:AF_debut}). Second, we describe the generation of a silver dataset (\autoref{sec:train_generation}), starting with the feature vector generation process for each patient identified as having an AF onset in the previous step (\ref{sec:vector_generation}), from now on AF-debut patients and, based on the patient-specific vectors of patients with an AF debut, we explain the process of detecting AF recurrence using multiple data sources (\ref{sec:recurrence_detection}), along with the characteristics of the resulting dataset (\autoref{sec:rec_data}).

\subsection{Cohort Generation: AF onset detection} \label{sec:AF_debut}
To study AF recurrence, it is essential to identify the subjects who meet the inclusion criteria, in this case, patients who have experienced AF onset. An AF onset is defined as the first documented episode of AF in a patient with no prior history of the condition. 

The inclusion process involves a series of steps aimed at identifying patients with AF onset through cross-verification of both structured and unstructured data. As noted in the introduction, structured data often contain errors and missing values. To address these issues, we propose a dual verification approach utilizing unstructured information:  

\begin{enumerate}
    \item \textbf{Structured information based patient selection} (Patient Filter): Identify patients with a codified AF onset according to the OBI Business Intelligence system (see  \autoref{subsec:patient_filter}).  
    \item \textbf{Unstructured information based patient selection for double checking} (Report Filter):  
    \begin{enumerate}
        \item \textit{LM Selector}: Retrieve discharge reports associated with the selected patients and perform an initial screening for potential AF onset cases. This step utilizes a fine-tuned EriBERTa-base Language Model (LM) \citep{eriBERTa}, specifically trained to detect AF onset mentions in clinical text.
        \item \textit{Regex-Based Selector}: Apply an additional checking step to refine data quality. This step employs a NLP and regular expression-based AF onset detection tool, described in detail in \autoref{subsec:report_filter}.
    \end{enumerate}
\end{enumerate}

This double checking methodology can be adapted to other healthcare systems depending on the quality of their structured data. Applying this methodology a final
accuracy of 97.81\% is obtained. Further details on the methodology and the underlying functions of the filtering modules can be found in \autoref{app:AF_debut}.

As a result of this process, a cohort of 2,861 patients was identified, each having an AF onset codified in the OBI system and documented in the collection of discharge reports.

\subsection{Silver training dataset Generation} \label{sec:train_generation}

The generation of the silver training dataset requires two main steps: first, extracting a tabular vector of relevant information for each hospitalization episode and, second, annotating each episode as either positive or negative for AF recurrence. Both structured and unstructured data are incorporated in these two steps.

\subsubsection{Tabular Data Generation} \label{sec:vector_generation}

A cohort of patients with an AF onset and the corresponding debut dates serves as the basis for automatically generating the tabular data, sometimes referred to as vectors in this study, required for subsequent AF recurrence prediction. 

\begin{figure*}[ht]
    \centering
    \includegraphics[width=15cm]{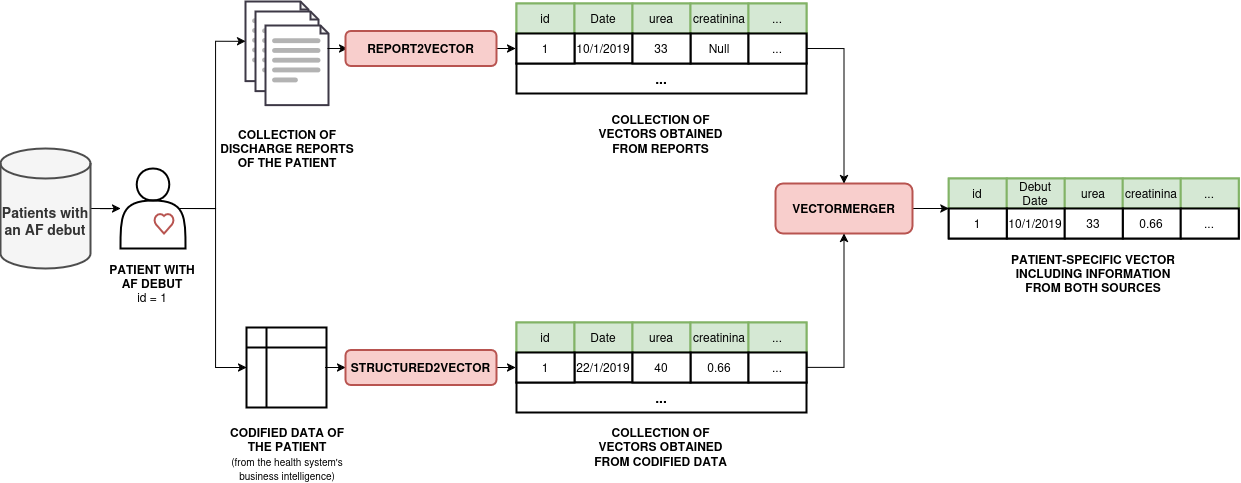}
    \caption{Summary of the vector generation process. Information from both free-text discharge reports and structured codified data from OBI system are merged in a single vector per patient describing its health status around the time of AF onset.}
    \label{fig:IG}
\end{figure*}

For each patient in the cohort, information is extracted from both unstructured and structured data. The process is divided into two steps, encompassing the extraction of information from both sources and the subsequent merging of the extracted data (see \autoref{fig:IG}):

\begin{enumerate}
    \item \textbf{Extract information}:
    \begin{itemize}
        \item \textit{Report2Vector}: Extracts relevant information from discharge reports and converts it into a tabular data format.
        \item \textit{Structured2Vector}: Extracts codified information from the OBI system and formats it into tabular data.
    \end{itemize}
    \item \textbf{Merge Information} (\textit{VectorMerger}): Combines the data extracted from both sources into a unified tabular format, where each row corresponds to a single patient.
\end{enumerate}

Additional details on each tool can be found in \autoref{app:vector_generation}. We must stress that the vector generation methodology proposed in this study can be adapted to other healthcare systems by modifying it to align with the specific data structure and resources available.

The generated tabular dataset includes features that comprehensively describe patients' health status, encompassing clinical data, laboratory results, and other relevant variables (see \autoref{app:features}). The proposed methodology, which integrates both structured and unstructured information, not only enhances data quality but also enables the extraction of clinically significant variables that are not codified in the system. For instance, left atrial parameters, such as size and volume, are well-established predictors of AF recurrence \citep{njoku2018left, shin2008left}. However, this information is often not codified in structured EHR data but can be found in free-text discharge reports. By leveraging the proposed text-processing methodology, key echocardiographic parameters, including left atrial size and left ventricular ejection fraction, were automatically extracted for the training set, while the test set was manually annotated by the cardiology department at BioBizkaia to ensure high-quality validation.

\subsubsection{Automatic labelling process: AF Recurrence Detection} \label{sec:recurrence_detection}

The annotation of AF recurrence is complex and subtle as the definition itself varies from studies. For this study the cardiology unit defined it as any clinical AF event from a month to 2 years after the AF onset. The cardiology department of BioBizkaia was in charge of annotating the test dataset.

A similar double-checking methodology, as used for identifying patients meeting the inclusion criteria, was applied to automatically label the training set with a Boolean indicator for AF recurrence in each patient.

The AF recurrence detection process operates as follows:
\begin{enumerate}
    \item For each patient, retrieve all its discharge reports from one month to two years following the initial AF onset.
    \item Generate a vector for each report using the \textit{Report2Vector} tool.
    \item Using the \textit{VectorMerge} tool, search through the vectors from one month to two years after the debut for the potential AF recurrence flag to assign the AF recurrence label for each patient:
    \begin{itemize}
        \item \texttt{true}: If evidence of a new AF episode is found in the \textit{Diagnostic} or \textit{Complementary Tests} sections.
        \item \texttt{false}: if there is no evidence of AF, but there is evidence of a return to sinus rhythm or an electrocardiogram (ECG) showing no AF diagnosis.
        \item Discarded patients: Cases where neither evidence of recurrence nor confirmation of sinus rhythm or non-AF ECG results is found.
    \end{itemize}
\end{enumerate}

The decision to exclude certain patients was based on the recognition that the absence of documented AF recurrence does not necessarily indicate that recurrence has not occurred. Many patients may not have undergone a follow-up electrocardiogram since their initial AF episode, or a subsequent electrocardiogram confirming a return to sinus rhythm may be unavailable. Given this uncertainty, a more restrictive selection approach was deemed appropriate to minimize the risk of misclassifying false negatives. To assess the reliability of the dataset, the quality of these annotations was evaluated by comparing them to a manually annotated test set. This assessment demonstrated that the annotation strategy achieved an accuracy of 83\%, indicating a high level of agreement with expert clinical evaluations.

This process results in 1,508 patients being labelled with AF recurrence between one month and two years after their debut episode. 

\subsection{Resulting AF recurrence dataset} \label{sec:rec_data}
The final characteristics of the dataset are visible in \autoref{tab:dataset}. Note that the percentage of recurrence across all the resulting subsets is quite high (around 63\%), which is coherent with the AF recurrence prevalence. AF recurrence and progression have been widely explored in recent studies \citep{vitolo2023factors,benali2023recurrences}. It should be noted that the progression of AF from paroxysmal to more sustained forms can be estimated around 30\% annually \citep{mcintyre2023atrial}. Besides, AF recurrence at 2 years even after pulmonary vein ablation is around 45\% in recent registries \citep{benali2023recurrences}, in which paroxysmal forms\footnote{Type of AF characterized by sudden, recurrent episodes of irregular heartbeats that resolve spontaneously within seven days.} are predominant. Nevertheless, in clinical practice, many first AF episodes (the ones included in this study) are not paroxysmal, and both persistent and permanent forms are commonly found at AF onset. Persistent types\footnote{Type of AF that lasts longer than seven days and requires medical intervention to restore normal rhythm.} are more likely to recur \citep{crowley2024catheter} and permanent forms\footnote{Type of long-term AF where the irregular heart rhythm is continuous and cannot be restored to normal, even with treatment.} are included in this study as AF onsets that recur (as factors conditioning AF recurrence and persistence will be analyzed in any forms of AF onsets). Taking all this information into consideration, an AF recurrence/persistence rate of 63\%, which can be found in our dataset, seems clinically plausible.

\begin{table}[!htb]
    \centering
    \begin{tabular}{l r r c}
        \toprule
        & \textbf{Size} & \textbf{\% Recurrence} & \textbf{Manual}\\ \midrule
        \textit{Train} & 1200 & 64.33 & No\\
        \textit{Test} & 308 & 62.33 & Yes\\ \bottomrule
    \end{tabular}
    \caption{Characteristics of the dataset. The leftmost two columns indicate the number of patient-specific vectors that comprise each subset, the middle one the percentage of AF recurrences within each subset and the rightmost one the annotation type.}
    \label{tab:dataset}
\end{table}

Before initiating the experimentation, a subset of patients was excluded from the dataset to minimize noise and ensure the reliability of the analysis. The excluded patients included those who died within three months of their AF onset episode and those aged over 90 years. This decision was based on the low clinical utility of AF recurrence over 90 years old, as rhythm control is anecdotic at this age and mechanisms of AF progression might vary. Besides, the high AF incidence in the context of multiorganic failure and last days of life could lead to inaccurate selection of features related to AF recurrence in this context, which might not be applicable in a more generalizable clinical context of a first-onset AF with further follow-up. Consequently, the final training set consisted of 978 patients, while the test set was reduced to 278 patients.

\section{Experimental setup}\label{sec:exp}

The experiments in this study aim to evaluate the predictive capacity of different models (traditional and tabular based models) for AF recurrence prediction 1 month to 2 years after the initial diagnosis using tabular data composed by clinical variables extracted from both structured and free-text data sources. In addition, the experiments aim to compare if the proposed models can deal with the challenges usually found in medical tasks, such as data scarcity, missing values, and outliers compared to current clinical scores.

All scripts and code used to carry out the experiments, including data preprocessing, model training, evaluation, and analysis, are publicly available on the following link: \url{https://github.com/anegda/PRAFAI-Vectors}.

This section is structured as follows. We first introduce the clinical scores used by clinicians (see subsection \ref{subsec:clinicalScores}), then we present the traditional machine learning learning techniques employed (see subsection \ref{subsec:MLModels}), and finally large tabular models (see subsection \ref{subsec:Large Tabular Model}).

\subsection{Clinical scores}
\label{subsec:clinicalScores}
 
The clinical scores selected for comparison in this study are CHADS2-VASc \citep{lip2010refining}, HATCH \citep{de2010progression}, and APPLE \citep{kornej2015apple}. These scores were chosen based on their established relevance in the prediction of AF onset and recurrence. CHADS2-VASc is widely used for stroke risk stratification in patients with AF, while HATCH has been proposed as a predictor of new-onset AF. Similarly, the APPLE score has demonstrated utility in assessing the likelihood of AF recurrence following catheter ablation. Given their clinical significance and widespread adoption, these scores serve as a meaningful baseline for evaluating the performance of our proposed model in predicting AF recurrence.

Since these tools generate numerical scores rather than direct classifications, it is necessary to define a threshold to transform them into a binary classification. The threshold values were determined by the cardiologist collaborator from BioBizkaia, following established guidelines and mainly literature where the scores were used in this setting. 

For the CHA2DS2-VASc score, a threshold of $\geq 2$ was chosen to indicate AF recurrence, based on the findings of \citet{vitali2019cha2ds2}. Similarly, the threshold for the HATCH score was set at $\geq 2$, following the recommendations in its original development study \citep{de2010progression}. Finally, the APPLE score also adopted a threshold of $\geq 2$, as established in its original validation study \citep{kornej2015apple}. 

The calculation of the clinical scores was performed using the generated tabular data. In cases where data were missing, the mode of the respective variable was imputed to maintain consistency and completeness in the dataset. The formulas for each score were applied as specified in their respective original publications, ensuring methodological accuracy and reproducibility.  

\subsection{ML models}
\label{subsec:MLModels}

The experimentation phase involved multiple ML techniques to address two primary challenges commonly encountered in medical datasets: missing values and high feature dimensionality. Initially, six model architectures were considered; however, after preliminary experiments, the selection was refined to three models: Support Vector Machine (SVM), Logistic Regression (LR), and XGBoost (XGB). This selection was based on their performance and the diversity of underlying learning principles, ensuring a comprehensive evaluation of different modeling approaches. 

The SVM and LR models were trained using the scikit-learn\footnote{\url{https://scikit-learn.org/stable/}} Python library, while the XGB model was implemented with the XGBoost\footnote{\url{https://xgboost.readthedocs.io/en/stable/index.html}} library. Feature selection and standardization techniques were also performed using scikit-learn, whereas sampling techniques were applied using the imbalanced-learn (\emph{imblearn})\footnote{\url{https://imbalanced-learn.org/stable/}} library.

\subsubsection{Handling Missing Values}

Missing data posed a significant issue in the generated dataset. Although XGB inherently handles missing values, preliminary experiments indicated that imputation was necessary for optimal performance across all models. Several imputation strategies were evaluated, with median imputation being selected due to its slightly better performance. An alternative approach involved training a logistic regression model to predict missing values based on existing data; however, this method was not viable due to the high proportion of missing entries, which led to poor model performance.

Moreover, the tabular numerical data was standardized using \textit{StandardScaler}, as it ensures a zero mean and unit variance, which benefits models sensitive to feature scaling. This choice was made after comparing it to alternative scaling methods, such as \textit{MinMaxScaler} and \textit{MaxAbsScaler}, which did not yield significant improvements in model performance. Standardization was therefore selected as the most effective and consistent preprocessing strategy.

\subsubsection{Feature Selection and Dimensionality Reduction}

The high dimensionality of the dataset complicated pattern extraction, particularly for simpler ML models. Initial experiments without feature selection yielded an MCC value (see \autoref{sec:metrics}) of 0 for SVM and LR, and a very low MCC of 0.04 for XGB. Analysis of the confusion matrices revealed a strong bias toward the majority class. While undersampling improved MCC, overall accuracy remained low (approximately 0.5) when all original features were included. Consequently, two feature selection techniques were applied: Recursive Feature Elimination (RFE) and LSFM (Lasso + SelectFromModel).
\begin{itemize}
    \item RFE: This method iteratively removes the least important features. A linear SVM was used as the estimator, ranking feature importance based on model coefficients and sequentially eliminating the least relevant variables.
    \item LSFM: This method applies Lasso regression (L1 penalty), which shrinks some feature coefficients to zero, effectively eliminating them. The absolute values of the remaining coefficients indicate feature importance, and SelectFromModel (SFM) then retains only the top 25\% most relevant features.
\end{itemize}

\subsubsection{Sampling Strategies}

Given the class imbalance in the dataset, sampling techniques were also considered. Although advanced methods such as SMOTE \citep{chawla2002smote} and TomekLinks \citep{elhassan2016classification} were tested, undersampling was ultimately selected. This approach involved removing rows with the highest proportion of missing values until the distribution between AF recurrence and non-recurrence was balanced at 50-50.

\subsubsection{Model Optimization and Validation}

To ensure robust performance, hyperparameter optimization was applied to each model. A 5-fold cross-validation strategy was implemented to prevent overfitting and eliminate the need for a separate development set. This approach ensured that models were trained and evaluated on multiple data splits, leading to more reliable performance estimates.

\subsection{Large Tabular Model (LTM)}
\label{subsec:Large Tabular Model}

In this study, we employ a tabular foundation model, also referred to as a LTM \citep{van2024tabular}. More concretely, we use TabPFN \citep{hollmann2025accurate} as the primary model for AF recurrence prediction due to its ability to efficiently handle complex tabular data while leveraging prior knowledge from pretraining.

Our task presents two main challenges related to the characteristics of the dataset; first, the limited number of patients available in the dataset; second, the high number of features in the vectors.Traditional deep learning architectures struggle with these constraints, whereas TabPFN has been specifically designed for small tabular datasets by learning from a broad range of simulated problems, making it well-suited for our use case.

Selecting TabPFN over other architectures such as TabNet \citep{arik2021tabnet}, AutoGluon \citep{erickson2020autogluon}, FT-Transformer \citep{gorishniy2021ft-transformer}, and NODE \citep{popov2019node} is justified by its superior performance in scenarios with limited data and high-dimensional feature spaces. Unlike traditional deep learning models that require extensive training, hyperparameter tuning, and large datasets, TabPFN is a prior-data fitted network that leverages knowledge from millions of synthetic tabular datasets, enabling it to generalize effectively even with small sample sizes. Moreover, since TabPFN does not require hyperparameter optimization, there is no need for a development set or a cross-validation strategy, which would otherwise reduce the size of the training or test sets.

\subsubsection{Dataset variants in LTM experimentation}
To comprehensively assess the model's capabilities, we conduct experiments using two variations of the dataset: (1) the preprocessed dataset used for training the ML models (TabPFN-pre, in Table \ref{tab:af_prediction}), where missing values were imputed and features standardized to improve consistency, and (2) the raw dataset with no prior imputation or transformations (TabPFN-raw, in Table \ref{tab:af_prediction}). This dual approach allows us to evaluate the extent to which the LTM can inherently address the challenges posed by missing values and feature scaling without requiring extensive preprocessing. 

\subsection{Evaluation Metrics}
\label{sec:metrics}

The performance of the models was evaluated using several standard classification metrics, including accuracy, precision, specificity and F1-score. 

Additionally, MCC is included as a more robust metric, particularly given the presence of class imbalance in the dataset. Unlike accuracy and ROC-AUC\footnote{ROC-AUC, is generated including predictions that obtained insufficient sensitivity and specificity, and does not reflect positive predictive value (precision) nor negative class predictive value (NPV).  
}, a high MCC value consistently reflects high values for all four fundamental rates derived from the confusion matrix (sensitivity, specificity, precision, and negative class predictive value), providing a more reliable and robust assessment of model performance across both classes (\cite{chicco2023matthews}). Its values range from -1, indicating complete misclassification, to 1, representing perfect classification, with 0 suggesting no predictive power beyond random chance. 
Furthermore, ROC-AUC is used to evaluate the model’s ability to distinguish between classes across various decision thresholds, with higher values indicating better overall discrimination. This metric was included as it is commonly used in similar studies as a primary benchmark for model evaluation, ensuring comparability with existing research.

\section{Results and Discussion}
This section begins by presenting the preliminary results of the ML models and the optimal data preprocessing strategy selected for further experimentation (see \autoref{sec:results_1}). It then examines the AF recurrence prediction performance of the developed models and compares them to traditional clinical scores using the general dataset (see \autoref{sec:results_2}). Additionally, gender and age bias analyses are explored in \autoref{sec:results_3} and \autoref{sec:results_4}, assessing potential disparities in model performance across different patient subgroups. Finally, a comparative analysis with similar studies is provided in \autoref{sec:results_5}.

\subsection{Results of ML models} \label{sec:results_1}
\autoref{tab:af_prediction_ml} presents the performance metrics for each ML model along with their optimal preprocessing strategies in the AF recurrence prediction task. While all models achieved the same accuracy (0.64), the SVM model obtained the highest MCC (0.16), followed by XGB (0.14) and LR (0.13). 
\begin{table}[h]
    \centering
    \renewcommand{\arraystretch}{1.2}
    \begin{tabular}{cccccc}
        \toprule
        \textbf{Model} & \textbf{MVI} & \textbf{FS} & \textbf{Sampling} & \textbf{ACC} & \textbf{MCC} \\
        \midrule
        SVM & Median & RFE & No & 0.64 & 0.16 \\
        LR & Median & RFE & No & 0.64 & 0.13 \\
        XGB & Median & LSFM & No & 0.64 & 0.14 \\
        \bottomrule
    \end{tabular}
    \caption{The best-performing machine learning architectures in the AF recurrence prediction task—Support Vector Machine (SVM), Logistic Regression (LR), and XGBoost (XGB)—along with their optimal missing value imputation (MVI), feature selection (FS), and sampling strategies. Performance is evaluated using accuracy (ACC) and Matthews Correlation Coefficient (MCC).}
    \label{tab:af_prediction_ml}
\end{table}

These experiments underscored the importance of feature selection algorithms in enhancing the learning process for traditional ML models. Models trained on the full set of original features exhibited limited predictive capacity, with a strong bias toward the positive class. This issue persisted even after applying an undersampling strategy, which, while balancing class distribution, resulted in reduced overall accuracy. Feature selection proved essential in mitigating these challenges, improving model performance by identifying the most relevant predictive variables.

\subsection{Results in AF recurrence prediction} \label{sec:results_2}

\autoref{tab:af_prediction} shows the final results of AF recurrence prediction, comparing traditional clinical scores, the SVM model, and the TabPFN in its two variations.

\begin{table*}[h]
    \centering
    \renewcommand{\arraystretch}{1.2}
    \begin{tabular}{llccccccc}
        \toprule
        & & \textbf{ACC} & \textbf{PRE} & \textbf{REC} & \textbf{F1} & \textbf{SPE} & \textbf{MCC} & \textbf{AUC} \\
        \midrule
        \multirow{3}{*}{Clinical Scores}& CHADS2-VASc & 0.6043 & 0.5363 & 0.6043 & 0.5378 & 0.1200 & -0.0052 & - \\
        & HATCH & 0.6043 & 0.5789 & 0.6043 & 0.5847 & 0.3000 & 0.0832 & - \\
        & APPLE & 0.4820 & 0.5671 & 0.4820 & 0.4809 & \textbf{0.6800} & 0.0510 & - \\ \midrule
        ML & SVM & 0.6367 & 0.6860 & \underline{0.7978} & 0.7377 & 0.3500 & 0.1626 & 0.6200 \\ \midrule
        \multirow{2}{*}{LTM} & TabPFN-pre & \textbf{0.6463} & \textbf{0.6943} & \textbf{0.7996} & \textbf{0.7433} & \underline{0.3733} & \textbf{0.1886} & \underline{0.6490} \\
        & TabPFN-raw & \underline{0.6415} & \underline{0.6917} & 0.7940 & \underline{0.7393} & 0.3700 & \underline{0.1784} & \textbf{0.6522} \\
        \bottomrule
    \end{tabular}
    \caption{Performance comparison of different models and clinical scores for AF recurrence prediction 1 month to 2 years after the debut episode. The metrics compared are Accuracy (ACC), Precision (PRE), Recall (REC), F1-score (F1), Specificity (SPE), Matthews Correlation Coefficient (MCC) and Area under the curve (AUC). The best performance in each metric is highlighted in bold, while the second-best is underlined in each column.}
    \label{tab:af_prediction}
\end{table*}

In the context of AF recurrence prediction in the general set, the key observations derived from the experimental results are as follows:  
\begin{itemize}
    \item TabPFN-pre achieved the highest overall performance. However, the performance of TabPFN-raw remains comparable, indicating that data preprocessing has not a measurable impact on the final results in this experiment\footnote{Paired t-test with p=0.0417}.
    \item The SVM model outperformed the clinical scores; however, achieving these results required an extensive data preprocessing pipeline, including missing value imputation and feature selection. In contrast, both variants of TabPFN also achieved similar performance without requiring such preprocessing.
    \item The difference in results between the SVM and TabPFN-pre (the best-performing model) is statistically significant\footnote{Paired t-test with p=0.0118}, indicating a better performance by TabPFN-pre. 
    \item The CHADS2-VASc and APPLE scores yield very low MCC values, close to what would be expected from random classification, indicating poor discriminative ability.  
    \item While CHADS2-VASc achieves a moderate accuracy, it exhibits a very low specificity, suggesting a significant difficulty in correctly identifying the negative class.  
    \item In contrast, the APPLE score achieves the highest specificity among the clinical scores. However, its overall accuracy remains below 50\%, limiting its practical usefulness. 
    \item The HATCH score outperforms the other clinical scores in terms of predictive performance, obtaining the highest MCC value among them (0.0832).
\end{itemize}

\subsection{Results for gender experiments} \label{sec:results_3}

Furthermore, to evaluate potential gender bias, an additional experiment was conducted by partitioning the training and test sets based on gender. The characteristics of these sets are detailed in \autoref{tab:dataset_gender}. Both gender-specific datasets were analyzed using the LTM model as well as the best-performing ML model to assess differences in predictive performance between genders.

\begin{table}[!htb]
    \centering
    \begin{tabular}{l c c c c c}
        \toprule
        & \multicolumn{2}{c}{\textbf{Train}} & & \multicolumn{2}{c}{\textbf{Test}} \\
        & Size & \% Recur & & Size & \% Recur \\ \midrule
        \textit{Male} & 505 & 60.79 &  & 131 & 64.12 \\
        \textit{Female} & 473 & 68,49 &  & 147 & 63.94 \\ \bottomrule
    \end{tabular}
    \caption{Characteristics of the datasets divided by gender for both train and test subsets.}
    \label{tab:dataset_gender}
\end{table}

In these experiments, both TabPFN variations were used, as the SVM model failed to achieve an MCC above 0. The results were compared to the HATCH and APPLE clinical scores, while the CHADS2-VASc score was excluded due to its inclusion of gender as a scoring variable. The results for both the female and male datasets are presented in \autoref{tab:af_prediction_gender}.

\begin{table*}[h]
    \centering
    \renewcommand{\arraystretch}{1.2}
    \begin{tabular}{llcccc}
        \toprule
        & & \textit{SET} & \textbf{ACC} & \textbf{F1} & \textbf{MCC} \\
        \midrule
        \multirow{4}{*}{LTM} & \multirow{2}{*}{TabPFN-pre} & \textit{Male} & \textbf{0.5954} & \textbf{0.7006} & \textbf{0.0830} \\
        & & \textit{Female} & \cellcolor{gray!25} \textbf{0.6621} & \cellcolor{gray!25} \textbf{0.7874} & 0.1747 \\ \cmidrule{2-6}
        & \multirow{2}{*}{TabPFN-raw} & \textit{Male} & 0.5929 & 0.6993 & 0.0757 \\
        & & \textit{Female} & 0.6621 & 0.7868 & 0.1729 \\ \midrule
        \multirow{4}{*}{Clinical Scores} & \multirow{2}{*}{HATCH} & \textit{Male} & 0.5496 & 0.5372 & -0.0222 \\
        & & \textit{Female} & 0.6530 & 0.6276 & \cellcolor{gray!25}\textbf{0.1875} \\ \cmidrule{2-6}
        & \multirow{2}{*}{APPLE} & \textit{Male} & 0.4656 & 0.4668 & 0.0070 \\
        & & \textit{Female} & 0.4966 & 0.4935 & 0.0905 \\
        \bottomrule
    \end{tabular}

    \caption{Performance comparison of the LTM models and traditional clinical scores for male and female datasets. The metrics compared are Accuracy (ACC), F1-score (F1) and Matthews Correlation Coefficient (MCC). To see the table with all the metrics see \autoref{app:af_prediction_gender} in \autoref{app:results}. The best performance for each metric is highlighted in bold for the male subset across models, while the top results for the female subset are shown in bold with a gray background.}
    \label{tab:af_prediction_gender}
\end{table*}

Regarding the gender-based experiments the key conclusions are as follows:
\begin{itemize}
    \item The ML models did not achieve a MCC higher than 0. This suboptimal performance may be attributed to the reduced size of the training sets after gender-based division and the application of 5-fold cross-validation which further shrinks the available training size, which can limit the models' ability to generalize effectively.
    \item Once again, the results for TabPFN-pre and TabPFN-raw are not statistically significant for the female set\footnote{Paired t-test with p=0.7621}. However, a statistically significant difference is observed in the male subset\footnote{Paired t-test with p=0.0170}. This suggests that the chosen imputation and standardization methods may help mitigate the decline in performance observed in the male subset.
    \item Both TabPFN models demonstrated relatively stable performance for the female dataset, achieving results comparable to those obtained with the entire dataset. However, the specificity was notably lower, potentially due to discrepancies in the distribution between the training and test sets for females (see \autoref{tab:dataset_gender}). The HATCH score showed comparable results to TabPFN for the female dataset, with higher specificity and MCC, but lower accuracy.
    \item Both TabPFN and clinical scores demonstrated significantly lower performance on the male dataset, with the HATCH score experiencing a notable decline of nearly 19 points in the MCC metric. To investigate potential causes, differences in feature distributions between the male and female subsets were analyzed, with a particular focus on the scoring variables of HATCH\footnote{The HATCH score considers the presence of hypertension, age > 75, transient ischemic attack or stroke, chronic obstructive pulmonary disease, and heart failure.}. A key finding was that the age distribution varied significantly between the two groups. This difference was statistically confirmed using a Student’s t-test, indicating a substantial disparity in age distribution between male and female patients. Additionally, the average age of female patients in the dataset was nearly 80 years, while the average age of male patients was 72 years (see \autoref{fig:age_gender}). This difference in average age could be attributed to the findings of \cite{stock2012cardiovascular}, which state that the mean age of onset for cardiovascular diseases (CVD) is generally higher in women, and that younger women are more likely to remain undiagnosed and untreated. These disparities may explain the pronounced performance drop of the HATCH score in the male subset, as age is a key contributing factor in its calculation.

    \begin{figure}[ht]
    \centering
    \includegraphics[width=8cm]{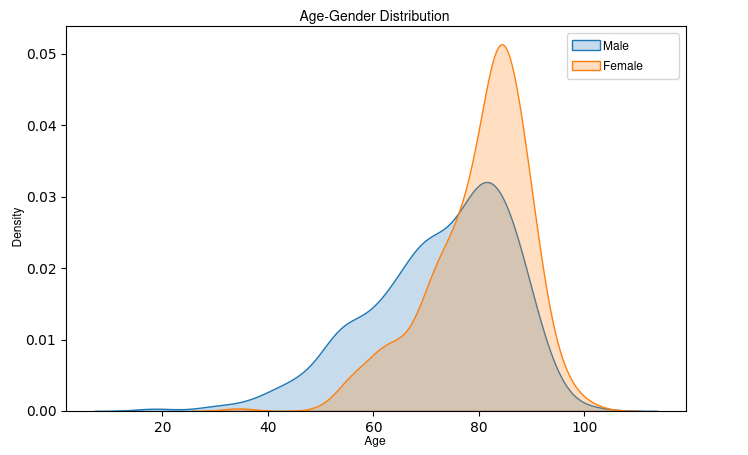}
    \caption{Age distribution comparison by gender, with females in orange and males in blue.}
    \label{fig:age_gender}
    \end{figure}

\end{itemize}

\subsection{Results for age experiments} \label{sec:results_4}
Given the observed differences in age distribution, an additional experiment was conducted on patients under the age of 75 to assess the performance of both the HATCH score and the TabPFN models in this specific subgroup. This experiment was based on the premise that the HATCH score performed worse in males due to the younger nature of this subset, potentially affecting its predictive reliability.  

Moreover, these experiments are particularly relevant, as rhythm control becomes less common at older ages, and the mechanisms of AF progression may differ. Consequently, accurately predicting AF recurrence within this age group is especially important for guiding therapeutic decisions and assessing clinical impact.

For consistency, both TabPFN models were utilized, as in previous experiments. The results of this analysis are presented in \autoref{tab:af_prediction_age}.
The main conclusions of these final experiments are the following:
\begin{itemize}
    \item The experiment confirms the strong dependency of the HATCH score on the age-based scoring feature and its reduced effectiveness in patients younger than 75 years old. While the decline in MCC appears moderate, the overall accuracy is notably low at 0.5478, highlighting the score’s limitations in this subgroup.
    \item Contrary to the previous experiments, this time the best metrics are encountered with the TabPFN-raw model, although this difference is not statistically significant\footnote{Paired t-test with p=0.8507}.
    \item Both TabPFN models demonstrate strong performance, achieving an MCC of 0.25 for TabPFN-raw and 0.27 for TabPFN-pre, which is double the MCC of the HATCH score, along with an accuracy of 0.66, the highest recorded across all experiments. These findings confirm that, with sufficient training data, the model effectively predicts AF recurrence in patients under 75 years old, outperforming traditional clinical scores.
    \item Once again, while the APPLE score achieves a moderate MCC, its accuracy remains below 0.5, indicating poor overall predictive reliability.
\end{itemize}

\begin{table*}[h]
    \centering
    \renewcommand{\arraystretch}{1.2}
    \begin{tabular}{llllccc}
        \toprule
        & & \textit{SET} & & \textbf{ACC} & \textbf{F1} & \textbf{MCC} \\ \midrule
        
        \multirow{6}{*}{LTM} & \multirow{3}{*}{TabPFN-pre} & \multirow{3}{*}{\textit{< 75}} & General & \underline{0.6613} & \underline{0.7364} & \textbf{0.2677} \\
        & & & \textit{Male} & 0.6410 & 0.7308 & 0.1932 \\
        & & & \textit{Female} & 0.6957 & 0.7470 & 0.3852 \\ \cmidrule{2-7}
    
        & \multirow{3}{*}{TabPFN-raw} & \multirow{3}{*}{\textit{< 75}} & General & \textbf{0.6638} & \textbf{0.7479} & \underline{0.2491} \\ 
        & & & \textit{Male} & 0.6438 & 0.7468 & 0.1483 \\
        & & & \textit{Female} & 0.6984 & 0.7503 & 0.3881 \\ \midrule
        
        \multirow{9}{*}{Clinical Scores} & \multirow{3}{*}{CHADS2-VASc} & \multirow{3}{*}{\textit{< 75}} & General & 0.5478 & 0.5357 & -0.0165 \\
        & & & \textit{Male} & 0.5616 & 0.5751 & 0.0736 \\
        & & & \textit{Female} & 0.5238 & 0.4120 & -0.0663 \\ \cmidrule{2-7}
        
         & \multirow{3}{*}{HATCH} & \multirow{3}{*}{\textit{< 75}} & General & 0.5478 & 0.5554 & 0.1216 \\
        & & & \textit{Male} & 0.5354 & 0.5507 & 0.0057 \\
        & & & \textit{Female} & 0.5714 & 0.5635 & 0.1783 \\ \cmidrule{2-7}
        
        & \multirow{3}{*}{APPLE} & \multirow{3}{*}{\textit{< 75}} & General & 0.4521 & 0.3875 & 0.1484 \\
        & & & \textit{Male} & 0.4383 & 0.3986 & 0.1615 \\ 
        & & & \textit{Female} & 0.4761 & 0.3650 & 0.0663 \\ \bottomrule
    \end{tabular}

    \caption{Performance comparison of the LTM models and traditional clinical scores for the  subset of patients younger than 75. The metrics compared are Accuracy (ACC), F1-score (F1) and Matthews Correlation Coefficient (MCC). For each model, the metrics obtained for the test set are presented, followed by two additional rows that display the metrics separately for the male and female subsets. To see the table with all the metrics see \autoref{app:af_prediction_age} in \autoref{app:results}.}
    \label{tab:af_prediction_age}
\end{table*}

\subsection{Comparison with other studies} \label{sec:results_5}

To the best of our knowledge, no prior research has specifically investigated AF recurrence within the same context as this study. As outlined in \autoref{sec:intro} and \autoref{sec:rel_work}, existing studies primarily focus on predicting the first occurrence of AF or assessing AF recurrence following catheter ablation. However, no study has been identified that examines AF recurrence occurring between one month and two years after the initial episode. Moreover, most studies on AF recurrence following catheter ablation focus on a one-year post-ablation period, whereas this study examines recurrence within a two-year window after the initial AF onset. This broader time frame provides a different perspective on AF progression and makes the task of prediction harder.

For instance, \cite{budzianowski2023machine} proposes a ML model designed to predict AF recurrence within the first year following catheter ablation. Their study utilized a relatively small dataset of 201 patients and incorporated 82 parameters derived from clinical, laboratory, and procedural variables. The best-performing model, based on the XGB architecture, achieved an F1-score of 0.547.

However, a direct comparison with our results is not entirely fair due to several key differences. On one hand, our study benefits from a larger training dataset, which generally enhances model performance. On the other hand, their patient selection criteria were more restrictive, potentially limiting the generalizability of their findings. Additionally, their analysis focused exclusively on AF recurrence within a one-year time frame, whereas our study considers a broader context. Despite these differences, our results demonstrate superior performance, with an F1-score that is 0.2 points higher. This improvement highlights the potential advantages of our approach, even when accounting for variations in dataset size and patient selection criteria.

\cite{roney2022predicting} developed 100 patient-specific models to simulate and predict AF recurrence in individuals who underwent their first ablation procedure. Their approach combined AF-related biophysical simulations, patient history metrics, and electrocardiogram data, which were processed using various traditional machine learning techniques. Among these, the SVM chieved the best performance, yielding an F1-score of 0.758. This result is notably close to the F1-score of 0.7433 reported in our current study.

Once again, a direct comparison between these results is not entirely appropriate due to several methodological differences. First, the patient selection criteria in \cite{roney2022predicting} differ from those in our study. Second, the recurrence time frame analyzed in their research is a year compared to the 2 years we include in the current study. Finally, their approach incorporates additional data types, such as biophysical simulations and ECG data, which were not utilized in our work. 

\section{Conclusions}

This study presents important advantages in the field of early diagnosis, particularly in the prediction of AF recurrence prediction, by addressing the two key research questions outlined in \autoref{sec:intro}.

The \textbf{RQ1} is addressed through the proposed automated dataset generation methodology. 

A major challenge in early diagnosis research is the incompleteness and potential errors in codified information from EHRs, a widespread issue across healthcare systems. Standard codified vectors often present errors and lack key clinical variables, limiting their usefulness in predictive modeling. To address this, the proposed methodology extracts valuable information from free-text discharge reports, enriching the dataset and significantly reducing missing values and errors.

To achieve this, the study introduces a systematic approach for converting unstructured clinical text into structured tabular data through a three-step NLP pipeline designed for AF recurrence detection. The quality of the tabular data describing the patients' health status and used for the prediction task underscores the importance of integrating complementary data sources in clinical studies. While this methodology is applied to AF recurrence in this study, it is highly adaptable and can be extended to other healthcare systems and disease prediction tasks, improving the quality and completeness of clinical datasets.

Moreover, in this study the combination of codified and free-text data enabled a silver standard annotation for AF recurrence, a variable not explicitly codified in the OBI system, bypassing the resource-intensive process of manual annotation and providing a scalable solution for data preparation.

Regarding \textbf{RQ2}, this study introduces models capable of predicting AF recurrence between one month and two years after the initial onset, providing a direct comparison with traditional clinical scoring systems. By focusing on this previously unexplored time frame, the study offers a new perspective on the early diagnosis of AF, extending beyond existing research that primarily focuses on predicting AF onset or AF recurrence within one year after therapeutic intervention. This broader predictive window enhances long-term risk assessment and, from a clinical standpoint, supports arrhythmologists in making informed treatment decisions. By identifying patients at risk of recurrence earlier, the proposed approach can help guide personalized treatment strategies, potentially improving patient outcomes and optimizing resource allocation in clinical practice.

Another key contribution of this study is the comparison between traditional clinical scoring systems and the proposed traditional ML and LTM architectures. The results obtained from the clinical scoring systems highlight the limitations of these approaches. While these scores offer interpretability and ease of use, their predictive capacity in this specific context remains insufficient. Among the evaluated clinical scores, HATCH demonstrated the best performance.

Both the traditional ML models and the LTM approach significantly outperformed the clinical scores for the whole patient cohort, underscoring the necessity of more sophisticated approaches that can leverage a broader set of predictive features and capture complex relationships between them. 

The LTM approach demonstrated robust predictive performance a la par with the SVM-based model, while also offering the advantage of requiring minimal data preprocessing. This confirms LTMs' ability to effectively handle missing values and high-dimensional data, mitigating the impact of noise without extensive preprocessing.

The results and model comparisons in this study also included a series of experiments designed to investigate potential gender and age biases in the predictive tools. These experiments aimed to evaluate whether the models performed equally well across different demographic subgroups, which is critical for ensuring fairness and generalizability in clinical applications.

The gender-based experiments revealed significant disparities in model performance between male and female cohorts. Notably, the male subset exhibited a substantial decline in performance across all methods, particularly for the HATCH scoring system. Further analysis of feature distributions highlighted a significant age disparity between genders, suggesting that age might be a confounding factor influencing the observed performance differences.

To address this, additional experiments were conducted focusing exclusively on patients under 75 years old. These experiments confirmed a strong dependency of the HATCH score on age, with performance varying considerably across age groups. In contrast, the proposed LTM approach demonstrated superior performance. As previously noted, the performance within this age group is highly significant, as the use of rhythm control strategies tends to decline with advancing age. Consequently, the limited effectiveness of traditional clinical scoring systems is particularly noteworthy, given the critical importance of accurately predicting AF recurrence in this population for guiding therapeutic decisions and evaluating clinical outcomes.

\section{Future Work}  

Thanks to the flexibility provided by the tabular data generation process described in \autoref{sec:vector_generation}, this study—while specifically focused on AF recurrence—can be extended to the early detection of other diseases. The dataset generation tools and regular expression-based methods already encompass various cardiovascular conditions. With appropriate modifications to the time frame criteria for vector overlap, these methodologies could be adapted to predict other disease occurrences and to other healthcare systems, broadening the applicability of this approach in clinical research.

Moreover, incorporating additional features and integrating traditional clinical scores as predictive inputs in an LTM model could be a promising direction for future research. This approach may enhance model performance by leveraging established clinical knowledge alongside data-driven learning.

In addition to improving dataset quality, future work will explore the development of novel methodologies for automatic annotation leveraging LLMs. Unlike the current approach, which relies on regular expressions and is inherently limited in its ability to generalize, LLM-based annotation systems could enhance the accuracy and scalability of the labelling process. These models have the potential to capture complex linguistic patterns and contextual information, leading to more precise and adaptable annotation strategies, ultimately improving the reliability of AF recurrence prediction.  

Beyond annotation improvements, further research could focus on developing multimodal predictive models that integrate not only codified EHR data and discharge reports but also additional clinical modalities such as electrocardiograms and physiological simulations. The incorporation of diverse data sources could provide a more comprehensive representation of patient conditions, potentially enhancing predictive performance by capturing multiple dimensions relevant to AF recurrence.  

Furthermore, transparency and interpretability are critical factors in the deployment of AI-driven models in clinical practice. While the LTM model has demonstrated strong predictive capabilities, its inherent ``black-box'' nature poses challenges for real-world medical applications. To address this, future research should incorporate explainability algorithms to provide insights into the model's decision-making process. Methods such as SHAP (Shapley Additive Explanations) and influence functions could be explored to identify the most relevant clinical variables contributing to AF recurrence predictions.

\section*{Acknowledgments}
This work has been partially supported by the HiTZ Center and the Basque Government, Spain (Research group funding IT1570-22) as well as by MCIN/AEI/10.13039/5011 00011033 Spanish Ministry of Universities, Science and Innovation  by means of the projects:
EDHIA PID2022-136522OB-C22 (also supported by FEDER, UE), DeepR3
TED2021-130295B-C31 (also supported by European Union NextGeneration
EU/PRTR).

A. G. Domingo-Aldama has been funded by ILENIA project 2022/TL22/00215335 and the Predoctoral  Training Program for Non-PhD Research Personnel grant of the Basque Government (PRE\_2024\_1\_0224).

A. García Olea has been funded by BioBizkaia grant under the code BB/I/PMIR/24/001
\bibliographystyle{cas-model2-names}
\bibliography{cas-refs}
\clearpage
\appendix
\onecolumn

\section{Dataset Features and Characteristics} \label{app:features}
\begin{longtable}{llll}\toprule
Characteristics &Type &Summary \\ \midrule
\textbf{Demographic data} & & \\
Gender &Binary &Female: 51.16\%, Male: 48,94\% \\
Age &Numeric &Mean (SD): 76.33 (12.25); Range: 18.0-103.0 \\
Pensioner &Binary &1: 85,32\%, 0: 14,68\% \\
Resident &Binary &1: 12.63\%, 0: 88,37\% \\
Weight &Numeric &Mean (SD): 75.02 (16.09); Range: 40.3-153.2 \\
Height &Numeric &Mean (SD): 1.61 (0.1); Range: 1.35-1.9 \\
IMC &Numeric &Mean (SD): 29.26 (5.26); Range: 16.4-58.04 \\
& & \\
\textbf{Laboratory results} & & \\
Urea &Numeric &Mean (SD): 54.1 (32.07); Range: 9.0-298.0 \\
Creatinine &Numeric &Mean (SD): 1.15 (0.8); Range: 0.15-9.65 \\
Albumin &Numeric &Mean (SD): 3.93 (0.5); Range: 1.52-7.8 \\
Glucose &Numeric &Mean (SD): 134.34 (56.56); Range: 48.0-824.0 \\
Hba1c &Numeric &Mean (SD): 6.27 (1.1); Range: 4.3-13.7 \\
Potassium &Numeric &Mean (SD): 4.29 (0.56); Range: 2.34-7.7 \\
Calcium &Numeric &Mean (SD): 9.35 (0.84); Range: 2.36-19.9 \\
HDL cholesterol &Numeric &Mean (SD): 49.85 (18.0); Range: 12.0-145.0 \\
LDL cholesterol &Numeric &Mean (SD): 117.92 (40.63); Range: 25.0-315.0 \\
no-HDL cholesterol &Numeric &Mean (SD): 93.34 (36.74); Range: 12.0-344.0 \\
Cholesterol &Numeric &Mean (SD): 170.6 (44.66); Range: 72.0-380.0 \\
Ntprobnp &Numeric &Mean (SD): 3489.18 (4869.44); Range: 59.34-35144.0 \\
Troponin TnT &Numeric &Mean (SD): 57.31 (174.72); Range: 5.0-2823.0 \\
Fibrinogen &Numeric &Mean (SD): 440.89 (156.38); Range: 120.0-1200.0 \\
Leuokcytes &Numeric &Mean (SD): 10.44 (23.92); Range: 1.4-500.0 \\
PCR &Numeric &Mean (SD): 4.08 (4.7); Range: 0.1-20.0 \\
TSH &Numeric &Mean (SD): 2.13 (1.63); Range: 0.0-13.52 \\
Sodium &Numeric &Mean (SD): 140.08 (3.97); Range: 104.0-165.0 \\
& & \\
\textbf{Procedures} & & \\
Echocardiogram &Binary &1: 11.64\%, 0: 88.36\% \\
Electrocardiogram &Binary &1: 50.17\%, 0: 49,83\% \\
LVEF &Numeric &Mean (SD): 54.89 (12.27); Range: -1.0-84.0 \\
Diameter left atrium &Numeric &Mean (SD): 42.04 (11.73); Range: 14.0-90.0 \\
Area left atrium &Numeric &Mean (SD): 29.02 (5.37); Range: 18.5-45.0 \\
Size left atrium &Categorical (c=4) & \\
& & \\
\textbf{Medical History} & & \\
Depression &Binary &1: 10.37\%, 0: 89.63\% \\
Alcohol &Binary &1: 14.68\%, 0: 85.32\% \\
Drug dependence &Binary &1: 00.21\%, 0: 99.79\% \\
Anxiety &Binary &1: 08.96\%, 0: 91.04\% \\
Dementia &Binary &1: 10.59\%, 0: 89.41\% \\
Renal insufficiency &Binary &1: 19.41\%, 0: 80.59\% \\
Menopause &Binary &1: 00.64\%, 0: 99.36\% \\
Osteoporosis &Binary &1: 09.53\%, 0: 90.47\% \\
Smoking &Binary &1: 32.32\%, 0: 67.68\% \\
SAHOS &Binary &1: 07.13\%, 0: 92.87\% \\
Hyperthyroidism &Binary &1: 04.30\%, 0: 95.7\% \\
EPOC &Binary &1: 13.76\%, 0: 86.24\% \\
Diabetes type 1 &Binary &1: 15.03\%, 0: 84.97\% \\
Diabetes type 2 &Binary &1: 23.92\%, 0: 76.08\% \\
Dyslipidemia &Binary &1: 26.89\%, 0: 73.11\% \\
Hypercholesterolemia &Binary &1: 37.26\%, 0: 62.74\% \\
Flutter &Binary &1: 19.76\%, 0: 80.24\% \\
Heart Failure &Binary &1: 51.94\%, 0: 48.06\% \\
Metabolic Syndrome &Binary &1: 00.21\%, 0: 99.79\% \\
Arterial Hypertension &Binary &1: 78.05\%, 0: 21.95\% \\
Ischemic cardiomyopathy &Binary &1: 18.70\%, 0: 81.3\% \\
Stroke &Binary &1: 13.34\%, 0: 86.66\% \\
Myocardiopathy &Binary &1: 07.62\%, 0: 92.38\% \\
Branch block &Binary &1: 18.21\%, 0: 81.79\% \\
Atrioventricular block &Binary &1: 05.72\%, 0: 94.28\% \\
Bradycardia &Binary &1: 06.35\%, 0: 93.65\% \\
Premature contractions &Binary &1: 14.18\%, 0: 85.82\% \\
Sinus node dysfunction &Binary &1: 00.99\%, 0: 99.01\% \\
Rheumatic valve disease &Binary &1: 00.28\%, 0: 99.72\% \\
Mitral Insufficiency &Categorical (c=5) &0: 81.23\%, 1: 14.40\%, 2: 02.75\%, 3: 01.20\%, 4: 00.42\% \\
Mitral Stenosis &Categorical (c=5) &0: 97.81\%, 1: 01.98\%, 2: 00.07\%, 3: 00.07\%, 4: 00.07\% \\
Aortic Stenosis &Categorical (c=5) &0: 89.77\%, 1: 09.10\%, 2: 00.35\%, 3: 00.28\%, 4: 00.49\% \\
Aortic Insufficiency &Categorical (c=4) &0: 84.69\%, 1: 14.68\%, 2: 00.42\%, 3: 00.21\% \\
Tricuspid insufficiency &Categorical (c=4) &0: 78.12\%, 1: 21.52\%, 2: 00.28\%, 3: 00.07\% \\
Peripheral Arteriopathy &Binary &1: 04.80\%, 0: 95.2\% \\
& & \\
\textbf{Drugs (ATC code)} & & \\
b01a &Binary &1: 97.67\%, 0: 02.33\% \\
n02ba01 &Binary &1: 01.91\%, 0: 98.09\% \\
a02bc &Binary &1: 89.34\%, 0: 10.66\% \\
c03 &Binary &1: 80.95\%, 0: 19.05\% \\
g03a &Binary &1: 00.07\%, 0: 99.93\% \\
a10 &Binary &1: 44.11\%, 0: 55.89\% \\
n06a &Binary &1: 56.67\%, 0: 43.33\% \\
n05a &Binary &1: 24.14\%, 0: 75.86\% \\
n05b &Binary &1: 57.87\%, 0: 42.13\% \\
c01 &Binary &1: 56.03\%, 0: 43.97\% \\
c02 &Binary &1: 11.71\%, 0: 88.29\% \\
c04 &Binary &1: 06.28\%, 0: 93.72\% \\
c07 &Binary &1: 62.88\%, 0: 37.12\% \\
c08 &Binary &1: 37.26\%, 0: 62.74\% \\
c09 &Binary &1: 67.61\%, 0: 32.39\% \\
c10 &Binary &1: 56.88\%, 0: 43.12\% \\
& & \\
\textbf{Atrial Fibrillation} & & \\
Type &Categorical (c = 4) &0: 79.96\%, 1: 17.57\%, 2: 0,99\%, 3: 17.57\% \\
Recurrence &Binary &1: 64.01\%, 0: 35.99\% \\
\bottomrule \\
\label{tab:dataset_char}
\end{longtable}

\clearpage
\section{Cohort Generation - detailed process} \label{app:AF_debut}

The methodology for identifying subjects who meet the inclusion criteria, specifically patients who have experienced AF onset, is detailed in \autoref{sec:AF_debut}. This section provides a more in-depth explanation of the filtering process, describing each module and the adaptations made to align with the available resources.

\subsection{Patient Filter} \label{subsec:patient_filter}

The initial step involves identifying patients who meet the inclusion criteria. Specifically, this includes individuals whose AF onset is recorded in the OBI system. This system comprises structured electronic health record (EHR) data, which includes the codified diagnosis, the date of diagnosis, and the patient ID. It is important to note that coding policies may vary over time and across different healthcare systems and hospitals. In some institutions, all diagnoses recorded in the patient file are coded and stored, whereas in others, only the primary diagnosis is encoded. Additionally, coding is prone to errors. Therefore, thorough verification and filtering are recommended to ensure accuracy.

\subsection{Report Filter} \label{subsec:report_filter}
The AF onset must be additionally documented using free-text in a discharge report from our designated cohort. An AF onset report is defined as a medical report that includes a reference of AF or related terms in the \textit{Diagnosis} and \textit{Complementary Tests} section of the discharge report, without any mention of prior AF episodes in the \textit{Medical History} section of that same report. 

A total of approximately 8,000 discharge reports are associated with patients who have codified AF onset according to the OBI system. To facilitate the identification of those discharge reports that corresponded to the onset of AF, two tools were developed: one tool based on NLP and regular expressions (\textit{Regex Filter}), and the other tool utilizing a LM (\textit{LM Filter}).

The \textit{Regex filter} applies a structured three-step process to meet the search criteria for AF onset detection:
\begin{enumerate}
    \item \textit{Section identification}: The tool published by \cite{section} was used for identifying the sections within the EHRs\footnote{More precisely, reason for consultation, past medical history, current disease, general exploration, complementary tests, diagnosis, treatment and evolution.}. Note that this step is crucial for fulfilling the search criteria described previously. The original article reports a F-score of 91.00\% for detecting \textit{Medical History}, 91.00\% for detecting \textit{Diagnosis} and 93.00\% for detecting \textit{Complementary Tests}.
    \item \textit{Medical Entity Recognition (MER)} and \textit{relation extraction}: The study conducted by \cite{unimer} addressed the MER task by identifying entities within the medical domain and establishing relationships between them. In our research, we utilized the DeepMER system, which is based on Recurrent Neural Networks (RNNs) for token classification. This system was trained on an annotated corpus of entities including six entity labels\footnote{qualifier, body structure, disease, drug, procedure, and allergy}. Specifically, we employed the drug label to identify medications used by patients. Additionally, we utilized the DeepNeg system \citep{unimer}, which is also based on RNNs, to detect instances of negation and reduce the occurrence of false positives. For example, it is common to encounter phrases such as "no diabetes mellitus" in a patient's medical history. In such cases, accurately identifying the negation is crucial to avoid incorrectly establishing the presence of the disease.
    \item \textit{Regular expression-based AF onset detection}: Leveraging the information gleaned from the preceding tools, the authors made use of a set of sophisticated regular expressions, developed in collaboration with a cardiologist, to detect AF-related terms, enabling the classification of EHRs as either indicative of an AF onset or not.
\end{enumerate}

Given the large volume of discharge reports to be processed (nearly 8,000), and the computational demands of the Regex Filter, which required a three-step process, a more efficient one-step filtering method was needed to streamline processing. As a result, an LM Filter was developed to significantly reduce the initial number of reports before applying the Regex Filter, optimizing computational efficiency. 
Furthermore, this two-step approach improves the reliability of AF onset detection by cross-validating outcomes from both filtering methods.  
The LM filter was designed as an efficient, single-step solution to minimize computational effort while improving generalizability. It also reduces the risk of error propagation associated with the modular structure of the Regex Filter, ensuring more robust and accurate predictions. Its main objective was to develop a binary classifier capable of identifying whether a given report referenced an AF onset episode. To accomplish this, a dataset comprising 5,944 instances for training, 743 for development, and 794 for testing was employed. To optimize efficiency and minimize manual effort, the training dataset was automatically annotated using the NLP and regular expression tool, resulting in a "silver" dataset. In contrast, the development and testing datasets were manually annotated by a team of five experts—four with a computer science background and one with expertise in cardiology. The inter-annotator agreement, measured using Fleiss' Kappa, was 0.9393, indicating near-perfect agreement. Among the five selected LM architectures, the best-performing model was based on the EriBERTa-base architecture \citep{eriBERTa}, achieving a final accuracy of 97.81\%.

\clearpage
\section{Tabular Data Generation - detailed process} \label{app:vector_generation}

The tabular data for the AF onset cohort is generated by extracting information from both discharge reports and codified EHR data, which are then integrated into a single row per patient, as detailed in subsection \ref{sec:vector_generation}.  

This process consists of three key steps: (1) extraction of information from discharge reports, (2) extraction of structured data from the OBI system, and (3) merging of both sources into a unified dataset. This section provides a detailed explanation of each step in the data integration pipeline.

\subsection{Report2Vector: Extracting information from discharge reports}
Information from discharge reports was integrated into the final AF recurrence dataset using a custom-developed tool named \textit{Report2Vector}. This tool transforms each discharge report into a row of tabular data (referred to as a report-specific vector) that contains relevant variables. These variables were identified in collaboration with a cardiologist from BioBizkaia, ensuring the inclusion of clinically significant information.

Each report-specific vector comprises a total of 86 variables, including the id of the patient and date of the report, demographic data, laboratory test results, and procedures (e.g., electrocardiograms, echocardiograms) along with their results. Additionally, the vector captures the patient's medical history, encompassing both cardiovascular and non-cardiovascular conditions, treatments, and four AF-related flags: whether a new AF diagnosis was made in the report, whether a previous AF episode is mentioned in the medical history, the type of AF the patient experienced, and whether the report may refer to a potential AF recurrence\footnote{AF recurrence cannot be identified from a single report, as it requires consideration of previous episodes and specific time intervals.}. Additional details about the data extracted from the vectors can be found in \autoref{tab:dataset_char}.

Ensuring accurate recall of variables is crucial not only for maintaining the quality of the dataset, but also for two other reasons. The first one is to properly supplement the information present in the OBI system, as certain key variables—such as echocardiogram results or AF recurrence status are not tracked by this system. Minimizing the number of missing information will result in a more complete dataset after the information from the OBI and free-text reports is merged. The second one is the correct detection of AF recurrence labels within the text, which will be later used as the silver labels for the training and validation datasets.

The tool is composed of the same modules described for the \textit{Regex filter}: a \textit{Section Identification} module, a \textit{MER and relation extraction} module, and a set of advanced \textit{Regular Expressions}. These regular expressions were developed in collaboration with the BioBizkaia cardiologist, who is actively involved in the project. %\footnote{The regular expressions are available at \url{https://github.com/anegda/PRAFAI-Vectors/blob/main/regular_expressions.json}} 

\subsection{Structured2Vector: Extracting structured information}
Information extraction from the OBI system was performed using the \textit{Structured2Vector} tool. This process is straightforward due to the use of standardized codes (ICD-10, ATC codes, etc.), which minimizes the variability commonly associated with free-text sources.

All available variables were converted into the same tabular format as that generated by the \textit{Report2Vector} tool. In this step, preserving the codification date of each variable is crucial to ensure the accurate merging of information in subsequent stages.

\subsection{VectorMerger: Merging information from both sources}
The tool responsible for merging vectors from both sources was named \textit{VectorMerger}. Although having the same format simplifies the merging of information from both sources, the process must still be performed meticulously, with careful attention to the appropriate time ranges for each variable of interest. Not all variables can be extracted from every vector, as each is linked to a specific temporal window. For example, past medical history can be retrieved from any prior vector, whereas laboratory results should be drawn only from vectors within a time frame of six months before to three months after the AF onset. These differences were established by the collaborator cardiologist from BioBizkaia to emulate the most similar laboratory findings at the AF onset.

The process of overlapping information begins by identifying the vector corresponding to the AF onset episode. Any missing information in this vector is subsequently filled using data from other available vectors. 

At this stage, AF recurrence detection is also performed, as identifying whether an AF episode is a recurrence requires knowledge of the patient's previous AF history. This process is explained in further detail in the next section (see subsection \ref{sec:recurrence_detection}).

\clearpage

\section{Results - detailed tables} \label{app:results}
\subsection{Results for gender experiments}

\begin{table}[pos=H]
    \centering
    \renewcommand{\arraystretch}{1.2}
    \begin{tabular}{llcccccccc}
        \toprule
        & & \textit{SET} &\textbf{ACC} & \textbf{PRE} & \textbf{REC} & \textbf{F1} & \textbf{SPE} & \textbf{MCC} & \textbf{AUC} \\
        \midrule
        \multirow{4}{*}{LTM} & \multirow{2}{*}{TabPFN-pre} & \textit{Male} & 0.5954 & 0.6667 & 0.7381 & 0.7006 & 0.3404 & 0.0830 & 0.5540 \\
        & & \textit{Female} & 0.6621 & 0.6587 & 0.9787 & 0.7874 & 0.1006 & 0.1747 & 0.7214 \\ \cmidrule{2-10}
        & \multirow{2}{*}{TabPFN-raw} & \textit{Male} & 0.5929 & 0.6643 & 0.7381 & 0.6993 & 0.3333 & 0.0757 & 0.5457 \\
        & & \textit{Female} & 0.6621 & 0.6595 & 0.9752 & 0.7868 & 0.1069 & 0.1729 & 0.7173 \\ \midrule
        \multirow{4}{*}{Clinical Scores} & \multirow{2}{*}{HATCH} & \textit{Male} & 0.5496 & 0.5295 & 0.5496 & 0.5372 & 0.2765 & -0.0222 & - \\
        & & \textit{Female} & 0.6530 & 0.6308 & 0.6530 & 0.6276 & 0.3208 & 0.1875 & - \\ \cmidrule{2-10}
        & \multirow{2}{*}{APPLE} & \textit{Male} & 0.4656 & 0.5438 & 0.4656 & 0.4668 & 0.6383 & 0.0070 & - \\
        & & \textit{Female} & 0.4966 & 0.5889 & 0.4965 & 0.4935 & 0.7169 & 0.0905 & - \\
        \bottomrule
    \end{tabular}
    \caption{Performance comparison of the TabPFN models and traditional clinical scores for male and female datasets. The metrics compared are Accuracy (ACC), Precision (PRE), Recall (REC), F1-score (F1), Specificity (SPE), Matthews Correlation Coefficient (MCC) and Area under the curve (AUC).}
    \label{app:af_prediction_gender}
\end{table}

\subsection{Results for age experiments}

\begin{table}[pos=H]
    \centering
    \renewcommand{\arraystretch}{1.2}
    \begin{tabular}{llllcccccccc}
        \toprule
        & & \textit{SET} & &\textbf{ACC} & \textbf{PRE} & \textbf{REC} & \textbf{F1} & \textbf{SPE} & \textbf{MCC} & \textbf{AUC} \\ \midrule
        
        \multirow{6}{*}{LTM} & \multirow{3}{*}{TabPFN-pre} & \multirow{3}{*}{\textit{< 75}} & General & \underline{0.6613} & \underline{0.7040} & \underline{0.7719} & \underline{0.7364} & 0.4861 & \textbf{0.2677} & \underline{0.6573} \\
        & & & \textit{Male} & 0.6410 & 0.7170 & 0.7451 & 0.7308 & 0.4444 & 0.1932 & 0.6020 \\
        & & & \textit{Female} & 0.6957 & 0.6814 & 0.8267 & 0.7470 & 0.5397 & 0.3852 & 0.7346 \\ \cmidrule{2-11}
    
        & \multirow{3}{*}{TabPFN-raw} & \multirow{3}{*}{\textit{< 75}} & General & \textbf{0.6638} & \textbf{0.7139} & \textbf{0.7854} & \textbf{0.7479} & 0.4524 & \underline{0.2491} & \textbf{0.6629} \\ 
        & & & \textit{Male} & 0.6438 & 0.7280 & 0.7667 & 0.7468 & 0.3768 & 0.1483 & 0.6249 \\
        & & & \textit{Female} & 0.6984 & 0.6875 & 0.8261 & 0.7503 & 0.5439 & 0.3881 & 0.7132 \\ \midrule
        
        \multirow{6}{*}{Clinical Scores} & \multirow{3}{*}{HATCH} & \multirow{3}{*}{\textit{< 75}} & General & 0.5478 & 0.5963 & 0.5478 & 0.5554 & \underline{0.6190} & 0.1216 & - \\
        & & & \textit{Male} &  0.5354 & 0.5947 & 0.5342 & 0.5507 & 0.5217 & 0.0057 & - \\
        & & & \textit{Female} & 0.5714 & 0.5996 & 0.5714 & 0.5635 & 0.7368 & 0.1783 & - \\ \cmidrule{2-11}
        
        & \multirow{3}{*}{APPLE} & \multirow{3}{*}{\textit{< 75}} & General & 0.4521 & 0.6596 & 0.4522 & 0.3875 & \textbf{0.9286} & 0.1484 & - \\
        & & & \textit{Male} & 0.4383 & 0.6898 & 0.4383 & 0.3986 & 0.9130 & 0.1615 & - \\ 
        & & & \textit{Female}& 0.4761 & 0.5738 & 0.4762 & 0.3650 & 0.9474 & 0.0663 & - \\ \bottomrule
    \end{tabular}

    \caption{Performance comparison of the TabPFN models and traditional clinical scores for the patient subset of younger than 75. The metrics compared are Accuracy (ACC), Precision (PRE), Recall (REC), F1-score (F1), Specificity (SPE), Matthews Correlation Coefficient (MCC) and Area under the curve (AUC). For each model, the metrics obtained for the test set are presented, followed by two additional rows that display the metrics separately for the male and female subsets.}
    \label{app:af_prediction_age}
\end{table}

\end{document}